\documentclass[two columns, final]{cvpr}

\usepackage{times}
\usepackage{epsfig}
\usepackage{graphicx}
\usepackage{amsmath}
\usepackage{amssymb}
\usepackage{algorithm}
\usepackage{algorithmic}
\usepackage{array}
\usepackage{booktabs}
\usepackage{makecell}
\usepackage{float}
\usepackage{diagbox}
\usepackage{changepage}
\usepackage{multirow}

\usepackage[pagebackref=true,breaklinks=true,colorlinks,bookmarks=false]{hyperref}

\def\confYear{CVPR 2021}

\begin{document}

\title{Spatial Feature Calibration and Temporal Fusion for Effective \\ One-stage Video Instance Segmentation}

\author{\textbf{Minghan Li}$^{1,2}$,\;  \textbf{Shuai Li}$^{1,2}$,\; \textbf{Lida Li}$^{1}$\; and \textbf{Lei Zhang}$^{1,2}$\footnotemark[1]\\
$^1$The HongKong Polytechnic University,\;  $^2$DAMO Academy, Alibaba Group\\
\textit{liminghan0330@gmail.com, \{ csshuaili,\, cslli,\, cslzhang\}@comp.polyu.edu.hk}\\
}



\maketitle
\pagestyle{empty}
\thispagestyle{empty}

\renewcommand{\thefootnote}{\fnsymbol{footnote}}
\footnotetext[1]{Corresponding author. This work is supported by the Hong Kong RGC RIF grant (R5001-18).}

\begin{abstract}
   Modern one-stage video instance segmentation networks suffer from two limitations. First, convolutional features are neither aligned with anchor boxes nor with ground-truth bounding boxes, reducing the mask sensitivity to spatial location. Second, a video is directly divided into individual frames for frame-level instance segmentation, ignoring the temporal correlation between adjacent frames. To address these issues, we propose a simple yet effective one-stage video instance segmentation framework by spatial calibration and temporal fusion, namely STMask. To ensure spatial feature calibration with ground-truth bounding boxes, we first predict regressed bounding boxes around ground-truth bounding boxes, and extract features from them for frame-level instance segmentation. To further explore temporal correlation among video frames, we aggregate a temporal fusion module to infer instance masks from each frame to its adjacent frames, which helps our framework to handle challenging videos such as motion blur, partial occlusion and unusual object-to-camera poses. Experiments on the YouTube-VIS valid set show that the proposed STMask with ResNet-50/-101 backbone obtains 33.5 \% / 36.8 \% mask AP, while achieving 28.6 / 23.4 FPS on video instance segmentation. The code is released online \href{https://github.com/MinghanLi/STMask}{https://github.com/MinghanLi/STMask}.
\end{abstract}

\section{Introduction}

Video instance segmentation aims to obtain the pixel-level segmentation mask for individual instances of all classes over the entire frames of a video, which heavily depends on spatial position-sensitive features to localize frame-level objects and redundant temporal information to track instances across frames. 
Following object detection and image instance segmentation works, 
modern video instance segmentation approaches usually adopt the top-down framework of first detecting and segmenting objects frame by frame and then linking instance masks across frames.


\begin{figure}
\begin{minipage}{0.48\linewidth}
  \centerline{\includegraphics[width=1\linewidth]{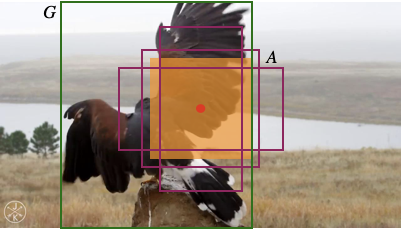}}
  \vspace{-1mm}
  \centerline{\small{(a) Original anchors and features\ }}
  \vspace{+1mm}
\end{minipage}
\hfill
\begin{minipage}{0.48\linewidth}
  \centerline{\includegraphics[width=1\linewidth]{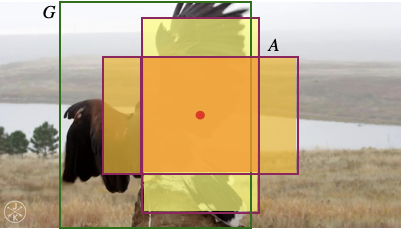}}
  \vspace{-1mm}
  \centerline{\small{(b) Calibrated anchors and features}}
  \vspace{+1mm}
\end{minipage}
\hfill
\begin{minipage}{0.48\linewidth}
  \centerline{\includegraphics[width=1\linewidth]{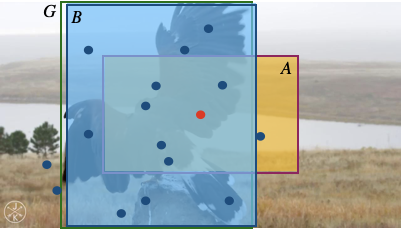}}
  \vspace{-1mm}
  \centerline{\small{(c) Adaptive features}}
  \vspace{+1mm}
\end{minipage}
\hfill
\begin{minipage}{0.48\linewidth}
  \centerline{\includegraphics[width=1\linewidth]{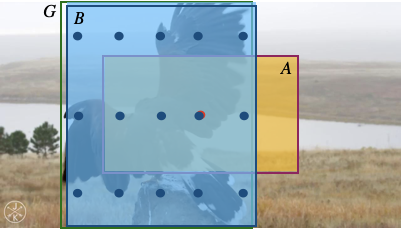}}
  \vspace{-1mm}
  \centerline{\small{(d) Aligned features}}
  \vspace{+1mm}
\end{minipage}
\caption{
Spatial calibration for anchor boxes and bounding boxes. 
(a) and (b) display anchors and features in original and calibrated one-stage networks respectively. 
(c) and (d) demonstrate adaptive and aligned features extracted from predicted bounding boxes.
Purple, blue and green rectangles denote anchors, predicted and ground-truth bounding boxes respectively, where coloured areas indicate the receptive filed of their convolutional features. 
}
\label{fig:FC}
\vspace{-3mm}
\end{figure}

Top-down video instance segmentation approaches can be divided into two-stage and one-stage methods. 
By adding a tracking branch to Mask R-CNN \cite{he2017mask}, two-stage video instance segmentation methods \cite{yang2019video,bertasius2020classifying,feng2019dual} first predict region-of-interests (RoIs) around ground-truth bounding boxes, and then feed aligned features via RoIPooling \cite{ren2015faster} or RoIAlign \cite{he2017mask} to segment frame-level object masks and to track cross-frame instances.
To obtain better location-sensitive features for mask predictor, many spatial feature calibration strategies for RoIs have been continuously proposed in recent years like Deformable RoI \cite{dai2017deformable} and Hybrid Task Cascade \cite{chen2019hybrid}.
For temporal information exploration, recently proposed MaskProp \cite{bertasius2020classifying} utilizes temporal features propagated from all frames of a video clip for clip-level instance tracking.
Obviously, two-stage methods have recognized the importance of spatial feature localisation and temporal feature tracking for video instance segmentation.
One-stage instance segmentation networks \cite{dai2016instance,xie2020polarmask,bolya2019yolact,tian2020conditional,cao2020sipmask,li2017fully,liang2017proposal}, which focus more on real-time speed, usually employ a fully convolutional network structure to directly predict the final mask for instances. 
Without RoIs of two-stage methods for localisation, early one-stage methods have to introduce extra position-sensitive information to improve segmentation performance like position-sensitive score maps \cite{dai2016instance} or semantic features \cite{chen2018masklab}.
In image domain, recently proposed anchor-based one-stage methods like Yolact \cite{bolya2019yolact} and CondInst \cite{tian2020conditional} decompose instance segmentation as the linear combination between instance-specific mask coefficients and instance-independent prototypes. 
Furthermore, SipMask \cite{cao2020sipmask}, introducing a tracking branch in Yolact\cite{bolya2019yolact}, achieves real-time speed but inferior performance in video instance segmentation task.

Analysing these anchor-based one-stage instance segmentation methods, we observe that, as shown in Fig.\ref{fig:FC} (a), multiple anchors of different shapes at each spatial position (purple rectangles) share same convolutional features (yellow area), which are neither aligned with pre-defined anchor boxes nor with ground-truth bounding boxes.
This fact does violate that instances segmentation is a spatial location-sensitive task. 
On the other hand, 
one-stage video instance segmentation methods directly divide a video into separate frames to perform image instance segmentation frame by frame and then track them across frames, which completely ignores high temporal correlation between adjacent frames.
This may fail to handle those challenging videos with motion blur, partial occlusion, or unconventional object-to-camera poses.
In other words, modern one-stage video instance segmentation methods achieve real-time speed at the cost of discarding spatial feature calibration and temporal feature correlation.

To address the issues, we propose a simple yet effective one-stage video instance segmentation framework, named STMask. 
Firstly, we design a feature calibration strategy for anchor boxes and ground-truth bounding boxes to obtain more precise spatial features.
Specifically, as shown in Fig.\ref{fig:FC} (b), to enable each anchor box can extract its own specific features, we first design multiple convolutional kernels at each spatial position, and then generate anchors according to the receptive field of these convolutional kernels.  
To improve feature presentation for objects segmentation and tracking, we first predict regressed bounding boxes around ground-truth bounding boxes by regression branch, and then extract features from them to segment and track instances. 
As shown in Fig.\ref{fig:FC} (c) and (d), we provide two strategies to extract features from regressed bounding boxes, including adaptive features by a single $1 \times 1$ convolutional layer and aligned features by mathematical derivation.  
Finally, we explore temporal correlation between video frames by adding a temporal fusion module to infer instance masks from adjacent frames, thereby improving the performance of objects detection, segmentation and tracking for those challenging videos.




\section{Related Work}
Video instance segmentation directly benefits from advances in image instance segmentation and video object detection field.
Thus this section consists of three parts.

\textbf{Image Instance Segmentation.}
Existing image instance segmentation methods either follow bottom-up or top-down paradigms. 
The bottom-up methods \cite{chen2018masklab,uhrig2018box2pix} widely employ multiple stages to first perform semantic segmentation and then identify the specific location of each instance by boundary detection \cite{kirillov2017instancecut}, pixel clustering \cite{liang2017proposal}, or pixels embedding loss \cite{neven2019instance,kong2018recurrent,newell2017associative}, position-sensitive pooling \cite{dai2016instance,li2017fully}. 
The top-down instance segmentation approaches first predict bounding boxes by object detectors and then perform mask segmentation within the predicted boxes. 
Mask R-CNN \cite{he2017mask} extends Faster R-CNN \cite{ren2015faster} by adding a mask segmentation branch on each Region of Interest (RoI), where RoIAlign operator is introduced to obtain `repool' features of each proposal for better mask prediction. 
Follow-up works try to improve its accuracy by aligning spatial features \cite{dai2017deformable}, enriching the FPN features \cite{liu2018path} or addressing the incompatibility between a mask’s confidence score and its localization accuracy \cite{huang2019mask}.
One-stage instance segmentation approaches \cite{xie2020polarmask,sofiiuk2019adaptis,chen2020blendmask,zhang2020mask,wang2019region,bolya2019yolact,wang2019solo,kirillov2020pointrend} recently are proposed to keep the trade-off between speed and performance.
Yolact-based methods \cite{bolya2019yolact,tian2020conditional,cao2020sipmask} break up instance segmentation into two parallel subtasks: generating a set of instance-independent prototypes and predicting instance-specific mask coefficients. 

\begin{figure*}[tbh!]
\begin{center}
\includegraphics[width=0.98\linewidth]{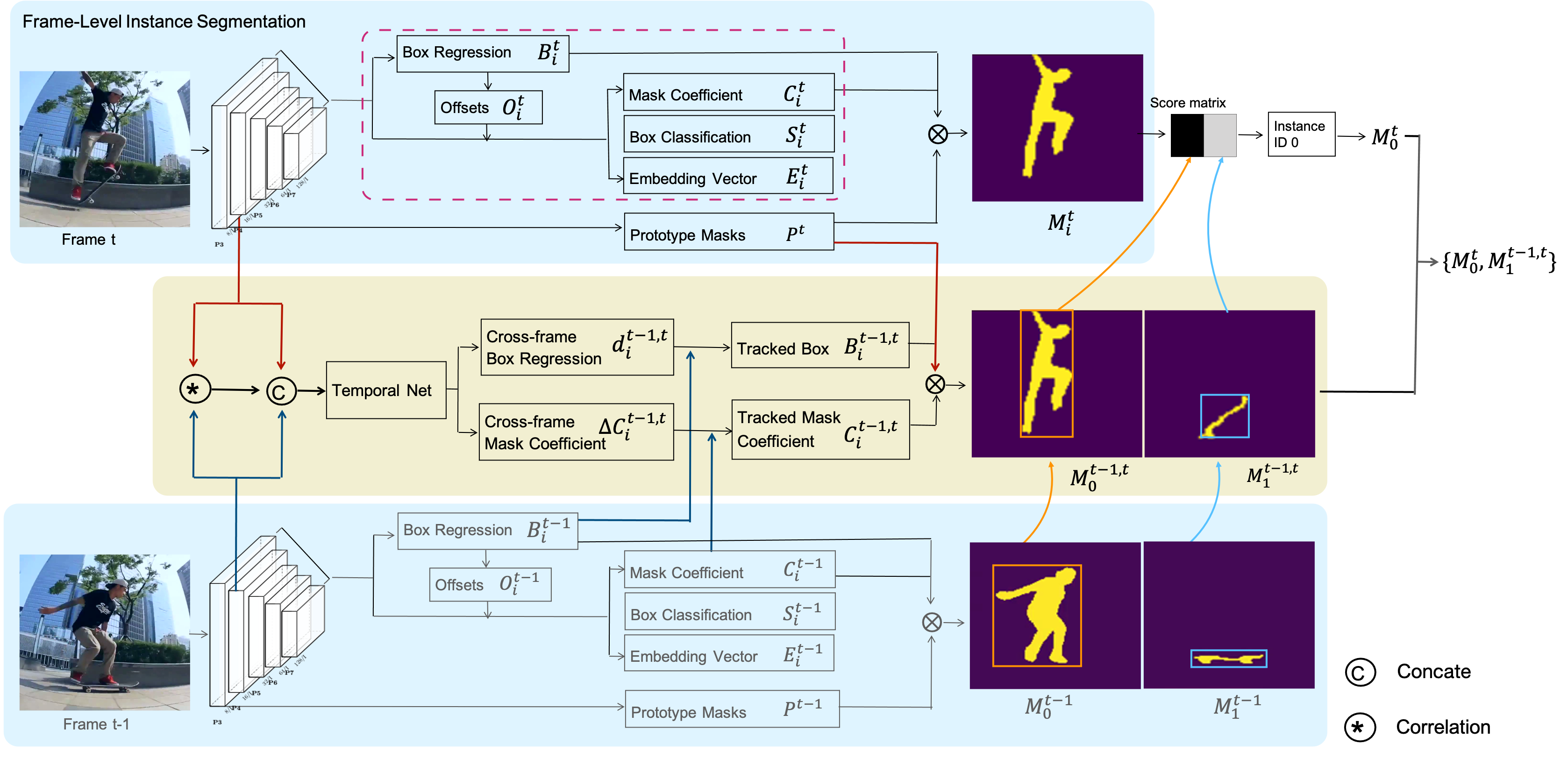}
\vspace{-3mm}
\end{center}
   \caption{An overview architecture of STMask, including frame-level instance segmentation (blue area) and cross-frame instance segmentation (yellow area).  
   Specifically, we first use Yolact with spatial calibration to obtain frame-level instance masks in time $t-1$ and $t$ respectively, then feed FPN features of two adjacent frames and the output of correlation operation to TemporalNet to infer the displacement of instances from time $t-1$ to $t$, and finally merge frame-level and cross-frame instance masks to obtain final instance masks.
   }
   \label{fig:overview}
   \vspace{-3mm}
\end{figure*}

\textbf{Video Object Detection.}
For handling challenging videos such as motion blur and occlusion, 
optical flow \cite{zhu2017flow,wang2018fully}, correlation operation \cite{feichtenhofer2017detect}, deformable convolutional networks \cite{bertasius2018object,xiao2018video} and relation networks \cite{deng2019relation,chen2020memory} are the popular technologies to propagate or align features across frames.
In addition, methods like\cite{deng2019object,shvets2019leveraging} try to utilise semantic similarity between frames to assist object detection in videos.

\textbf{Video Instance Segmentation.}
To jointly perform the detection, segmentation and tracking tasks simultaneously, most of video instance segmentation methods extend Mask R-CNN \cite{he2017mask} by adding a new branch for tracking.
For example, MaskTrack R-CNN \cite{yang2019video} predicts an extra embedding vector for each instance and use an external memory to store them for tracking across frames.
MaskProp \cite{bertasius2020classifying} introduces a mask propagation branch to propagate frame-level instance masks from each video frame to all the other frames in a video clip and then match clip-level instance masks for tracking, providing state-of-the-art instance segmentation performance and quite limited speed.
Besides, \cite{lin2020video} proposes a modified variation autoencoder (VAE) architecture built on the top of Mask R-CNN \cite{he2017mask}.
Recently, SipMask \cite{cao2020sipmask} also introduces a tracking branch (same as \cite{yang2019video}) in the one-stage image instance segmentation network Yolact \cite{bolya2019yolact} to obtain inferior performance but real-time speed. 
Besides, inspired by Guided Anchor \cite{wang2019region,li2019dynamic}, SipMask further aligns feature with regressed bounding boxes to improve feature representation for classification and mask coefficients generation.
Existing spatial feature calibration for one-stage approaches only focus on the misalignment between features and regressed bounding boxes. 


\section{STMask}
The overall architecture of STMask shown in the Fig. \ref{fig:overview} consists of frame-level instance segmentation with spatial calibration and cross-frame instance segmentation by temporal fusion module. 

\subsection{Spatial Calibration}

Our goal is to align features with anchors and ground-truth bounding boxes respectively for one-stage instance segmentation methods. 
For spatial feature calibration on anchors, multiple convolutional kernels shown in Fig. \ref{fig:FCA-arch} are introduced to alleviate the misalignment between features and anchors.
For spatial feature calibration on bounding boxes, shown in the pink rectangle of Fig. \ref{fig:overview}, we first individually predict regressed bounding boxes around ground-truth bounding boxes, and then extract features from those regressed bounding boxes to classify, segment and track instances.

\subsubsection{Feature Calibration for Anchors (FCA)}
\begin{figure}[!tb]
\begin{minipage}{1\linewidth}
  \centerline{\includegraphics[width=1\linewidth]{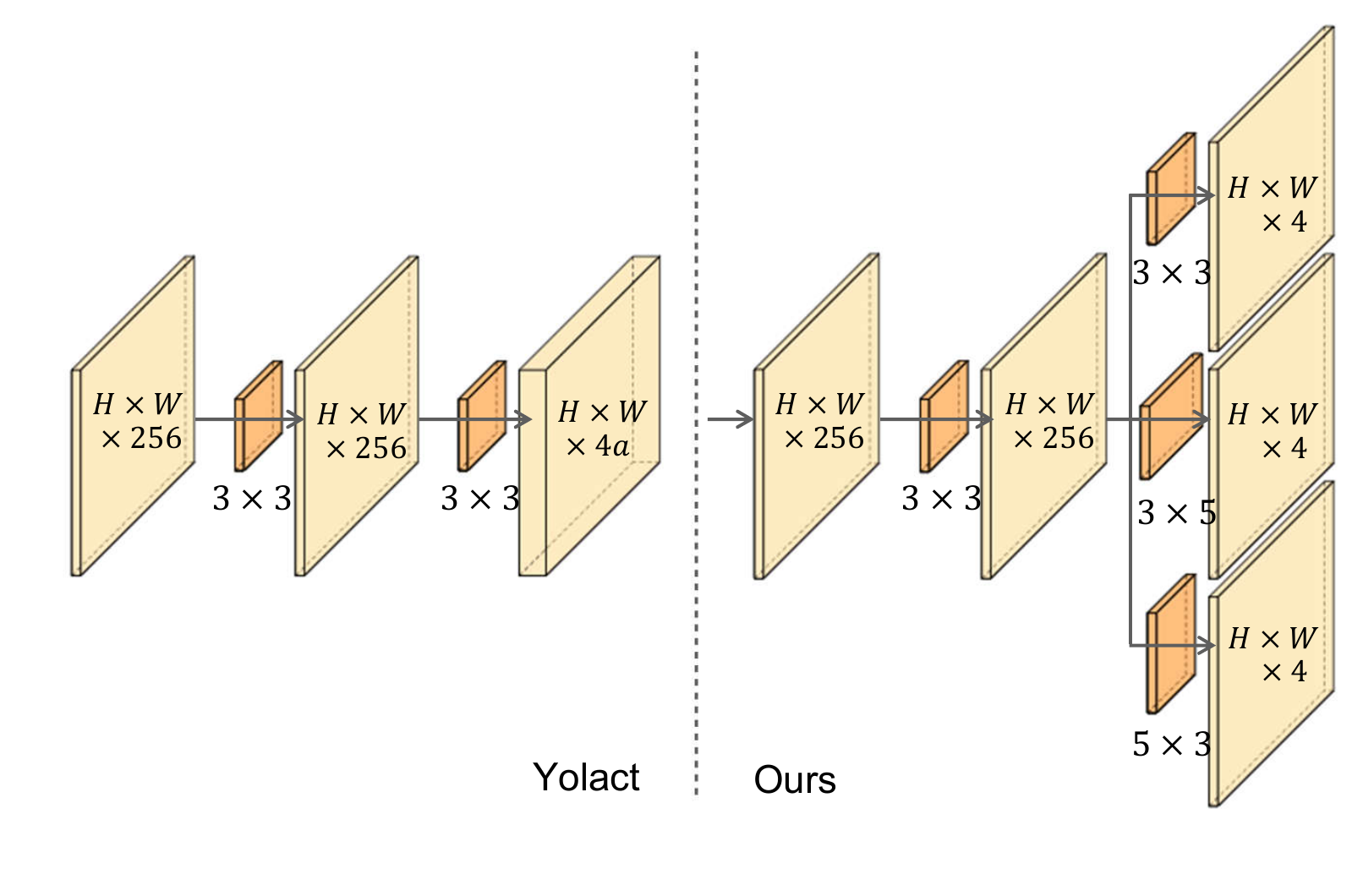}}
  \vspace{-2mm}
\end{minipage}
\caption{
Bounding Box Regression Architecture. \ Taking the regression branch as example, we design multiple convolutions for calibrating features with pre-defined anchor boxes, where $a\!=\!3$ is the number of anchors.
}
\label{fig:FCA-arch}
\vspace{-3mm}
\end{figure}

\vspace{-1mm}
One-stage anchor-based object detectors usually sample a large number of regions in the input image, determine whether these regions contain objects of interest, and adjust the edges of the regions so as to more accurately predict the ground-truth bounding boxes of the targets. 
In general, the sliding-window fashion to generate anchors is the most popular method, which generates multiple boxes with different scales and aspect ratios while centring on each pixel. 
As shown in Fig. \ref{fig:FC}(a), for multiple anchors with different shapes, one-stage detectors directly employ a $3\times 3$ convolution on the central point of anchors to extract features.
In practice, the receptive field of convolution should be positively related to the size of anchor boxes.
For example, those larger anchor boxes should have larger receptive fields, while those smaller boxes should have smaller ones.
To address this issue, thus, we adopt multiple convolution kernels with different aspect ratios on each FPN layer. 
For example, we replace the single $3\times 3$ convolution with three new aspect ratios of convolutions, $3\times 3,\ 3\times 5,\ 5\times 3$ respectively. The FCA architecture implemented on the bounding box regression branch is shown in Fig. \ref{fig:FCA-arch}. 
To further ensure the calibration between convolutional features and anchors, we keep the scales unchanged while changing the anchor aspect ratios from [1, 1/2, 2] to [1, 3/5, 5/3], which are same as the aspect ratios of convolutions. 
The simple feature calibration for anchors maintains position sensitivity for object detection and segmentation.

\subsubsection{Feature Calibration for Bounding Boxes (FCB)}
\vspace{-1mm}
After the feature calibration with anchors in last part, features for classification, mask coefficients and embedding vectors of tracking branch still fail to align with the regressed bounding boxes. 
To further address the issue, we adopt the prediction architecture of first predicting regressed bounding boxes and then extracting features from the regressed bounding boxes to segment and track objects.

Let $\{(P_i, G_i)\}_{i=1,...,N}$ denote the pairs of anchor boxes and ground-truth bounding boxes. Note that subscript $i$ is only added when necessary.
$P=(P_{x}, P_{y}, P_{w}, P_{h})$ specifies the pixel coordinates of the centre and the width and height of the anchor box. The same goes for ground-truth bounding box $G = (G_{x}, G_{y}, G_{w}, G_{h})$. 
The bounding-box regression aims to learn a transformation $d$ that maps an anchor box to its ground-truth bounding box:
\begin{align}\label{transformation}
\vspace{-1mm}
d=[d_{x}, d_{y}, d_{w}, d_h].
\end{align}
After that, its regressed bounding box $B_i$ can be calculated by applying the transformation\cite{girshick2014rich} 
\begin{align}\label{bbox_regression}
\vspace{-2mm}
   & B_x = P_wd_x + P_x,\quad\ \ B_y = P_hd_y + P_y; \nonumber \\
   & B_w = P_w\exp(d_w),\quad B_h = P_h\exp(d_h).
\end{align}
Based on the transformation between anchor boxes and predicted bounding boxes, we can introduce a 2D deformable convolution to calibrate convolutional features from anchor boxes to predicted bounding boxes.
For easier understanding, we take a $3\times 3$ convolution with dilation 1 as an example to explicate the process of spatial feature calibartion for bounding boxes. For each location $p_0$ on the output map $\mathbf{g}$, the 2D deformable convolution can be formulated as 
\begin{align}\label{dcn}
\mathbf{g}(p_0) = \sum_{p_n\in \mathcal{R}} w(p_n) \cdot \mathbf{f}(p_0+p_n+\Delta p_n),
\vspace{-1mm}
\end{align}
where $\Delta p_n$ is the offsets on the position $p_n \in \mathcal{R}$, which is a point of the grid  
\begin{equation}
\vspace{-1mm}
\mathcal{R} = \left\{
\begin{array}{lr}
    (-1, -1) \ (-1, 0) \ (-1, 1) \\
    (\ \ 0, -1) \ \ (\ \ 0,\ 0) \ \ (\ 0,\ 1) \\
    (\ \ 1, -1) \ \ (\ \ 1,\ 0) \ \ (\ 1,\ 1) \\
\end{array}
\right\},
\end{equation} 
To extract convolutional features from predicted bounding boxes, the offsets $\mathcal{O}$ should be dominated by the transforamtion $d$. Thus, this paper provides two strategies to obtain offsets: adding a single $1\times 1$ convolution layer to predict adptive offsets, or directly deriving the aligned offsets through mathematical geometric knowledge.

\begin{figure}[!tb]
\hfill
\begin{minipage}{0.49\linewidth}
  \centerline{\includegraphics[width=1\linewidth]{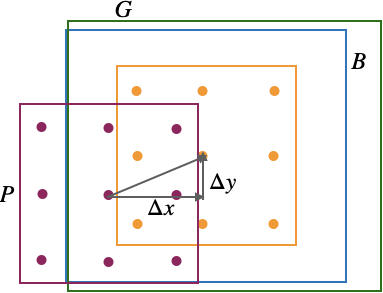}}
  \vspace{-1mm}
  \centerline{\small{\quad (a) Scale-invariant Translation}}
  \vspace{+1mm}
\end{minipage}
\hfill
\begin{minipage}{0.49\linewidth}
  \centerline{\includegraphics[width=0.97\linewidth]{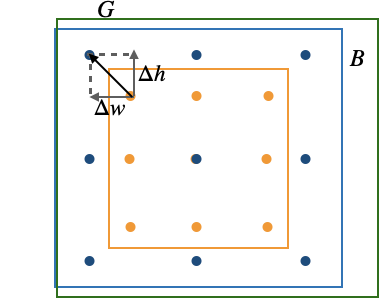}}
  \vspace{-1mm}
  \centerline{\small{\qquad (b) Scale Transformation }}
  \vspace{+1mm}
\end{minipage}
\caption{
Illustration of aligned features from anchor boxes to predicted bounding boxes. 
}
\label{fig:bbox_offset}
\vspace{-3mm}
\end{figure}

\textbf{Adaptive Features on Bounding Boxes.} \
Inspired by the anchor-guided feature adaption module \cite{wang2019region}, we also transform the feature at each individual location based on the underlying anchor transformation:
\begin{align}
\vspace{-2mm}
   \mathcal{O} = \mathcal{N}_O(d), 
\vspace{-1mm}
\end{align}
where $\mathcal{N}_O$ is a $1 \times 1$ convolutional layer to predict offsets according to the bounding-box regression transformation $d$. 
When the predicted offsets $\mathcal{O} \in R^{2\times 3 \times 3}$ is fed into a deformable convolutional layer to produce adaptive features shown in blue points in Fig. \ref{fig:FC}(c). 
For such regression-dependent offsets, each regressed bounding box can learn its own adaptive features to perform further objects classification, segmentation and tracking.

\textbf{Aligned Features on Bounding Boxes.} \
For the regression transformation $d$ in terms of four functions in Eq. \ref{transformation}, the first two specify a scale-invariant translation of the centre of anchor box, while the second two specify log-space translations of the width and height of anchor box. 
According to the four parameters of transformation $d$, the derivation process of generating the offsets can also be divided into two steps: \textit{scale-invariant translation} and \textit{scale transformation}, shown in Fig. \ref{fig:bbox_offset} (a) and (b) respectively.
On one hand, Fig. \ref{fig:bbox_offset} (a) demonstrates that all sampling points on the grid $\mathcal{R}$ of the anchor box have the same scale-invariant translation as the centre point. 
On the other hand, Fig. \ref{fig:bbox_offset} (b) shows that the absolute scale transformation of the width and height on the grid $\mathcal{R}$ is related to their own coordinate position.
Overall, the derived offsets for all points on the grid should be
\begin{equation}
\vspace{-1mm}
\mathcal{O} = (\Delta y, \Delta x) I + (\Delta h, \Delta w) \mathcal{R}.
\end{equation} 
where $I$ is a matrix with all elements as 1, $\Delta x, \Delta y, \Delta h$ and $\Delta w$ are listed below:
\begin{align}
\vspace{-1mm}
&\Delta x =k_wd_x, \ \Delta w = \exp(d_w)-1, \\
&\Delta y =k_hd_y, \ \Delta h\ = \exp(d_h)-1.
\vspace{-1mm}
\end{align}
where $k=(k_w, k_h)$ is the kernel size on the width and the height. Due to the limit of space, the detailed formula derivation process is provided in the \textbf{supplementary materials}.
In fact, the mathematical derivation for the offset in this part is equivalent to a special case of RoIAlign operation in two-stage networks, where each bin only takes the central point as its output value.

\subsection{Temporal Fusion Module}
Compared with image instance detection and segmentation, video instance segmentation task faces more challenges such as partial occlusion, unusual view, motion blur. 
To address the issue, thus, we further set up a temporal fusion module for cross-frame bounding boxes regression and mask segmentation.
The yellow area of Fig. \ref{fig:overview} shows the schematic diagram of temporal fusion module between two adjacent frames, which concatenate features of FPN between two adjacent frames to infer the displacement of instances from the previous frame to the current frame, thereby obtaining cross-frame instance masks from the previous frame to the current frame. Afterthat, we only need to merge frame-level detected instance masks and cross-frame tracked instance masks to get the final instances masks. Such a double guarantee architecture does improve the accuracy of detection and segmentation in video domain.

\textbf{Frame-level Instance Segmentation.}\ Same as Yolact \cite{bolya2019yolact}, STMask decomposes instance segmentation into the linear combination between instance-independent prototypes $P \in R^{H'\times W' \times k}$ and instance-specific mask coefficients $C \in R^{ k \times n}$.
Essentially, the process of learning prototypes $P$ is equivalent to learn the online basis set, also called dictionary \cite{mairal2009online}, when each instance $i$ can find a linear combination $C_i \in R^{ k \times 1} $ of a 'few' atoms from prototypes that is `close' to the ground-truth mask $M_i \in R^{ H' \times W'}$. The each instance segmentation can be implemented efficiently using a single matrix multiplication and the activation function (sigmoid function $\sigma$):
\begin{equation}\label{eq:instance_segmentation}
\vspace{-1mm}
M_i = \mathbf{Crop}(\sigma (PC_i),\ B_i). 
\end{equation} 
where the final masks $M_i$ is also cropped by the predicted bounding box $B_i$.

\begin{figure}[!tb]
\begin{minipage}{1\linewidth}
  \centerline{\includegraphics[width=0.96\linewidth]{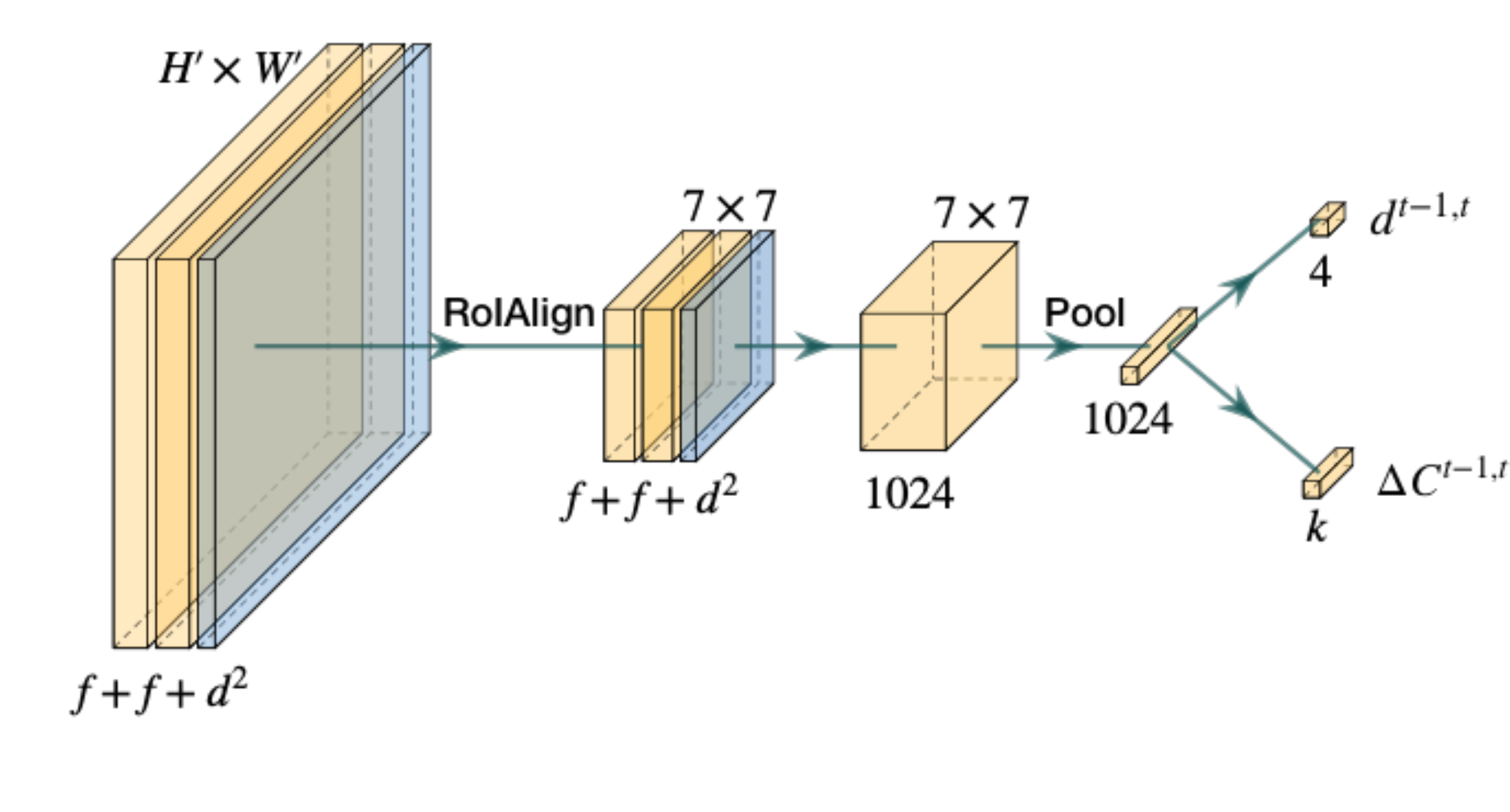}}
  \vspace{-3mm}
\end{minipage}
\caption{
Temporal Network Architecture. }
\label{fig:TF-arch}
\vspace{-4mm}
\end{figure}

\textbf{Cross-frame Instance Segmentation.}\ Given a pair of frames $I^{t-1}, I^{t} $ from a video $\mathcal{V} \in R^{H\times W \times V}$, the goal of proposed temporal fusion module is to track those object masks from time $t\!-\!1$ to $t$, noted as $M^{t-1, t}$.
Since the prototypes are instance-independent, according to Eq. (\ref{eq:instance_segmentation}), the temporal fusion module intuitively needs to predict the mask coefficients $C^{t-1, t}$ and bounding boxes $B^{t-1, t}$ of instances from time $t\!-\!1$to $t$. {}

Inspired by RoI tracking process in video object detection task \cite{feichtenhofer2017detect}, the temporal fusion module of STMask should include a bounding boxes regressor and a mask coefficient predictor.
To be specific, as shown in the yellow area of Fig.~\ref{fig:overview}, 
we first adopt a correlation operation on FPN features of $x^{t-1} \in R^{H'\times W' \times f}$ and $x^t \in R^{H'\times W' \times f}$ to embed motion information of instances, noted as $x_{corr}^{t-1, t} \in R^{H'\times W' \times d^2}$, where $d$ is the side length of local square.
Then the concatenation of features $\{x^{t-1}, x^{t}, x_{corr}^{t-1, t}\}$ is transported to the temporal network, shown in Fig. \ref{fig:TF-arch}, to infer the displacement of bounding boxes $d^{t-1, t}$ and mask coefficients $\Delta C^{t-1, t}$ of instances from time $t-1$ to $t$, where
\begin{align}
\vspace{-1mm}
d^{t-1, t} = \{d_{x}^{t-1,t}, d_{y}^{t-1,t}, d_{w}^{t-1,t}, d_{h}^{t-1,t}\}.
\end{align}
Thus, the cross-frame bounding boxes $B^{t-1, t}$ from time $t\!-\!1$ to $t$ can be formulated as
\begin{align}
\vspace{-1mm}
 &B_x^{t\!-\!1, t}\!=\! B_w^{t\!-\!1}d_x^{t\!-\!1,t}\!+\!B_x^{t-1},\ B_y^{t\!-\!1, t}\! =\! B_h^{t-1}d_y^{t\!-\!1,t}\!+\!B_y^{t-1}, \\ 
 &B_w^{t\!-\!1, t}\!=\! B_w^{t\!-\!1}exp(d_w^{t\!-\!1,t}), \ B_h^{t\!-\!1, t}\! =\! B_h^{t-1}exp(d_h^{t\!-\!1,t}). 
\end{align}
where the predicted bounding box coordinates $B^{t-1}$ acts as an anchor.
On the other hand, due to the similarity of the prototypes between adjacent frames, the mask coefficients of each instance on different frames should also be 'close' to each other. 
Based on the mask coefficients $C^{t-1}$ in the $t\!-\!1$ frame,
the cross-frame mask coefficients $C^{t-1, t}$ inferred from time $t\!-\!1$ to $t$ can be obtained by 
\begin{align}
 C^{t-1, t} = C^{t-1} + \Delta C^{t-1, t}.
\end{align}
Finally, the instance masks $M^{t-1, t}$ inferred from time $t\!-\!1$ to $t$ can be conducted by 
\begin{equation}\label{cross-frame-instance-seg}
\vspace{-1mm}
M^{t-1, t} =\mathbf{Crop}(\sigma (P^{t}C^{t-1, t}), B^{t-1, t}). 
\end{equation}

\textbf{Merge Frame-level and Cross-frame Instance Segmentation.} \ We denote the set of video-level instance IDs as $\mathcal{Y}$, which is incrementally built, and $N_t$ is the number of instances predicted by frame-level instance segmentation of Eq. (\ref{eq:instance_segmentation}) at time $t$.  
At the beginning, we assign IDs $\mathcal{Y}=\{1,\dots ,N_1\}$ for all frame-level object segmentations in the first frame of a video.
For following video frames, we first perform frame-level instance segmentations of Eq. (\ref{eq:instance_segmentation}) to predict frame-level instance masks $M^t$, and then execute the temporal fusion module of Eq. (\ref{cross-frame-instance-seg}) to infer cross-frame instance masks $M^{t-1, t}$ from time $t-1$ to $t$. 
Obviously, the cross-frame instance masks $M^{t-1, t}$ are naturally given the same embedding vectors $E^{t-1}$ and the same instance IDs as $M^{t-1}$.
Thus, the matching scores between frame-level masks $M^t$ and cross-frame masks $M^{t-1,t }$ consist of two components: the cosine similarity of embedding vectors and mask IoU:
\begin{align} 
s_{ij}^t = \alpha \frac{E^t_i \cdot E^{t-1}_j} {\parallel E^t_i\parallel \parallel E^{t-1}_j\parallel} + \beta\ \mbox{MIoU}(M^t_i, M^{t-1, t}_j)
\vspace{-2mm}
\end{align}
where $i \in \{1, \dots  N_t\}, j \in \mathcal{Y}$, and $\alpha, \beta$ are hyperparameters to balance the effect of different components. 
For each frame-level instance mask $M^t_i$, let $s_i^t = \max_{j \in  \mathcal{Y}} s_{ij}^t$ be the maximum score among all instances IDs $j \in  \mathcal{Y}$. If $s_i^t$ is greater than a certain threshold $\epsilon$, the frame-level instance mask $M^t_i$ will be assigned the instance ID with the highest score. Otherwise, it will be considered as a new one and will be assigned an instance ID $|\mathcal{Y}|+1$. 
Overall, the frame-level instance masks $M^t_i$ will be assigned instance IDs by:
\begin{equation} 
y_i^t = \left\{
\begin{array}{ll}
 \arg \max_{j \in  \mathcal{Y}} \ s_{ij}^t, \quad & \mbox{if}\ s_i > \epsilon, \\
|\mathcal{Y}| + 1,\quad  & \mbox{otherwise}. 
\end{array}
\right.
\end{equation}
We denote $\mathcal{Y}^t=\{y_i^t|i \in \{1, \dots  N_t\}\}$ as the set including all frame-level instance IDs at time $t$.
If there are instance IDs that do not appear in the set of frame-level instance IDs $\mathcal{Y}^t$ but appear in the set of cross-frame instance IDs $\mathcal{Y}$, 
in other words, $${\mathcal{Y}-\mathcal{Y}\cap \mathcal{Y}^t \neq \emptyset},$$
these cross-frame instance masks $M^{t-1, t}_{\mathcal{Y}-\mathcal{Y}\cap \mathcal{Y}^t}$ will be supplemented as the missing instances in the $t$ frame.
Finally, the merged masks of all instances in the $t$-th frame should be 
\begin{equation}
\bar{M}^t=M^{t} \cup M^{t-1, t}_{\mathcal{Y}-\mathcal{Y}\cap \mathcal{Y}^t}.
\end{equation}

\vspace{-1mm}
\section{Experimental Results}


\textbf{Training.} 
We conduct video instance segmentation experiments on YouTube-VIS \cite{yang2019video} dataset using the standard metrics.
In training, the YouTube-VIS train set is split as two subsets: train-sub and valid-sub set. One is used for training and the other is used as the valid set during training. 
We train all models on pairs of frames, where the second frame in a pair is randomly selected with a time gap $\delta \in $ [-5, 5] relative to the first frame, using pre-trained models on MS COCO dataset \cite{lin2014microsoft}.
Similar as MaskTrack R-CNN\cite{yang2019video}, we use the input size $360\times 640$ for training.
All model is trained with batch size 16 for 160k iterations and divide the learning rate at 66K and 133K. Training takes 1-2 days on four NVIDIA 2080Ti.

\textbf{Inference.}
Given an input image, we forward it through the frame-level network to obtain the outputs including classification scores, bounding box, embedding vectors, mask coefficients and the prototypes. 
Box-based Non-maximum suppression (NMS) with the threshold being 0.5 is used to remove duplicated detections and then the top 100 bounding-boxes are used to compute frame-level masks. 
After that, we forward the features of the previous frame and the current frames through temporal fusion module to first infer masks from the previous frame to current frame, then assig IDs for frame-level masks of the current frame and supplement the missing masks.
We take the hyperparameters as $\alpha=1, \beta=1$ in temporal fusion module.

\subsection{Design Choices}

In this section, we discuss the design choices of spatial calibration and temporal fusion module for our proposed one-stage video instance segmentation framework STMask. 
Similar as MaskTrack R-CNN \cite{yang2019video}, we extend Yolact \cite{bolya2019yolact} by adding a tracking branch to predict embedding vectors for tracking.
For this part, we adopt ResNet101 with deformable convolution layers (interval=3 ) as the backbone of Yolact \cite{bolya2019yolact}.
Note that all experiments do not include other improvements in Yolact++ \cite{bolya2019yolact++} such as more anchors and semantic segmentation loss.

\textbf{Design Choices of FCA.}
Our baseline Yolact \cite{bolya2019yolact} has three anchors with aspect ratios [1, 1/2, 2] and a $3 \times 3$ convolution in prediction head. 
As shown in Table \ref{tab:ablation_FCA} (1st row), it achieves 28.9\% in mask AP. 
After adding FCA strategy, the setting of aspect ratios with smaller convolutions $\{3 \times 3, 3 \times 2, 2 \times 3\}$ still brings the gain of 1\% mask AP, and the setting with larger convolutions of aspect ratios $\{3 \times 3, 3 \times 5, 5 \times 3\}$ brings a significant improvement of 1.9 \% mask AP.
Note that, for fair comparison, we set the kernel size to nearly $3 \times 3$ to ensure that the performance gain does come from feature calibration on anchors rather than a larger receptive field. For all experiments below, we choose the latter setting by default.

\textbf{Design choices of FCB.}
To analyse the impact of the feature calibration for regression bounding boxes (FCB) on classification, mask coefficients and embedding vectors of tracking branch, we discuss some possible combinations among them on the prediction head.
As shown in Table \ref{tab:ablation_FCB}, our baseline with FCA (1st row) achieves 30.8\% mask AP. 
Overall, all experiments adding FCB with adaptive and aligned features strategy on all possible combination among branches of the prediction head obtain significantly improved results on all metrics, around  1-3 \%.  
Besides, all experiments with aligned features calibration is relatively lower than that of adaptive features, because the latter may be more adaptable than the former.
By default, FCB (ada) and FCB (ali) denote FCB with adaptive features on all three branches, and that with aligned features on mask coefficient and tracking branches, respectively.

\begin{table}[!tb]
\caption{ Design choices for FCA on YouTube-VIS valid set.}
\vspace{-2mm}
\begin{center}
\setlength{\tabcolsep}{0.5mm}{
      \linespread{2}
      \begin{tabular}{p{0.12\textwidth}<{\centering}
      p{0.045\textwidth}<{\centering}p{0.048\textwidth}<{\centering}p{0.048\textwidth}<{\centering}
      p{0.045\textwidth}<{\centering}p{0.05\textwidth}<{\centering}p{0.045\textwidth}<{\centering}}
        \Xhline{1pt}
          Aspect ratios     &AP         &\ AP$_{50}$ &\ AP$_{75}$ &\ AP$_{S}$ &\ AP$_{M}$ &\ AP$_{L}$ \\
         \Xhline{0.5pt}

         \{3:3, 3:3, 3:3\}      & 28.9      & 45.8     & 30.1  & 9.8  & 26.9 & 40.3\\
         \{3:3, 3:2, 2:3\}      & 29.9      & 48.9     & \textbf{31.2}  & 8.0  & 27.8 & 40.3\\ 
         \{3:3, 3:5, 5:3\}      & \textbf{30.8}      & \textbf{49.8}     & 31.0  & \textbf{11.8} & \textbf{29.7} & \textbf{44.6}\\ 
         \Xhline{1pt}
      \end{tabular}\label{tab:ablation_FCA}
}
\vspace{-4mm}
\end{center}
\end{table}

\begin{table}[!tb]
\caption{ Design choices for FCB with adaptive and aligned features on YouTube-VIS valid set.}
\vspace{-2mm}
\begin{center}
\setlength{\tabcolsep}{0.5mm}{
      \begin{tabular}{p{0.12\textwidth}<{\centering}
      p{0.055\textwidth}<{\centering}p{0.06\textwidth}<{\centering}p{0.06\textwidth}<{\centering}
      p{0.075\textwidth}<{\centering}p{0.067\textwidth}<{\centering}}
         \Xhline{1pt}
         Baseline+FCA  & +Class     &  +Mask       & +Track     &\ AP(ada)   &\ AP(ali)\\
         \Xhline{0.5pt}
         \checkmark     &            &             &             &30.8       &30.8    \\ 
         \checkmark     & \checkmark & \checkmark   & \checkmark &33.7       &31.7    \\ 
         \Xhline{0.2pt}
         \checkmark     & \checkmark & \checkmark   &           &\textbf{34.4}      & 31.7 \\ 
         \checkmark     & \checkmark &              & \checkmark &34.3       &32.4 \\ 
         \checkmark     &            & \checkmark   & \checkmark &32.5     & \textbf{33.1}  \\ 
         \Xhline{1pt}
      \end{tabular}\label{tab:ablation_FCB}
}
\vspace{-4mm}
\end{center}
\end{table}

\textbf{Design Choices of Temporal Fusion Module.} \ 
We first conduct experiments on different FPN layers and the side length of local square for the correlation operation.
As shown in Table \ref{tab:TF-design}, the P4 FPN layer with the side length $11$ obtains the better performance for temporal fusion module.

\begin{table}[!tb]
\caption{ Design choices for temporal fusion module.}
\vspace{+1mm}
\begin{center}
\setlength{\tabcolsep}{0.5mm}{
      \linespread{2}
      \begin{tabular}{p{0.09\textwidth}<{\centering}p{0.04\textwidth}<{\centering}
      p{0.06\textwidth}<{\centering}p{0.055\textwidth}<{\centering}p{0.055\textwidth}<{\centering}
      p{0.055\textwidth}<{\centering}p{0.065\textwidth}<{\centering}}
        \Xhline{1pt}
         Layers         &  d           &\ AP  &\ AP$_{50}$ &\ AP$_{75}$ &\ AR$_1$ &\ AR$_{10}$  \\
         \Xhline{0.5pt}
         P3          &   13            &33.8 & 51.3 & 35.8 & 32.1 & 39.6 \\
         P4          &   11            &\textbf{35.0} & \textbf{53.0} & \textbf{38.7} & \textbf{32.8} & \textbf{40.1} \\ 
         \Xhline{1pt}
      \end{tabular}\label{tab:TF-design}
}
\vspace{-5mm}
\end{center}
\end{table}

\subsection{Ablation Study }

We conduct the ablation study on YouTube-VIS valid set using Yolact with ResNet101-DCN backbone\cite{bolya2019yolact++} as our basline. 
For fair comparison, the baseline of all experiments adopts three anchors and does not include semantic segmentation loss. 
Table \ref{tab:ablation_VIS_all} shows the impact of progressively integrating our different components: feature calibration with anchor boxes (FCA), adaptive / aligned feature calibration with anchor boxes (FCB), and temporal fusion module (TF) to the baseline. 
The four components progressively increase segmentation performance to 36.8 \% mask AP.
Besides, to better explore the individual improvements for mask segmentation brought by each single component, we also conduct the experiments adding the different components individually to the baseline shown in Table \ref{tab:ablation_VIS_single}, where FCA, FCB(ada), FCB(ali) and TF component respectively contribute to the gains of 1.9 \% , 2.3 \% , 2.4 \%  and 5.2 \%  mask AP.
Among these components, the FCB(ada) and TF provide the most improvement in accuracy over the baseline. 
These results suggest that each of our components individually contributes towards improving the final performance. 

We also perform more ablation study for STMask on COCO dataset to verify the effectiveness of spatial feature calibration on image instance segmentation task, which are provided in the \textbf{supplementary material}.

\begin{table}
\caption{ Ablation study on YouTube-VIS valid: progressively integrating different components into the baseline. }
\vspace{-2mm}
\begin{center}
\setlength{\tabcolsep}{0.5mm}{
      \linespread{2}
      \begin{tabular}{p{0.065\textwidth}<{\centering}p{0.055\textwidth}<{\centering}
      p{0.055\textwidth}<{\centering}p{0.05\textwidth}<{\centering}
      p{0.045\textwidth}<{\centering}
      p{0.05\textwidth}<{\centering}p{0.05\textwidth}<{\centering}p{0.05\textwidth}<{\centering}}
         \Xhline{1pt}
         Baseline       & +FCA       &  +FCB     & +TF    &\ AP   &\ AP$_{50}$ &\ AP$_{75}$ &FPS \\ 
         & \multicolumn{3}{c}{\ (ada\  /\ ali)} \\
         \Xhline{0.5pt}
         \checkmark     &            &              &   &28.9   & 45.8  & 30.1  & 31.2\\ 
         \checkmark     & \checkmark &              &   &30.8   & 49.8  & 31.0 & 29.5\\ 
         \Xhline{0.2pt}
         \checkmark     & \checkmark & \checkmark\ / &  &\textbf{34.4}   & \textbf{53.0}  & 34.6 & \textbf{28.3}\\ 
         \checkmark     & \checkmark & \quad  / \ \checkmark   &     & 33.1 & 51.4 &\textbf{35.2} & 26.7 \\
         \Xhline{0.2pt}
         \checkmark     & \checkmark & \checkmark\  /    &\checkmark &\textbf{36.8}   & \textbf{56.8}  & 38.0 & \textbf{23.4} \\ 
         \checkmark     & \checkmark & \quad  / \ \checkmark   &\checkmark &36.3   & 55.2  & \textbf{39.9} & 22.1\\ 
         \Xhline{1pt}
      \end{tabular}\label{tab:ablation_VIS_all}
}
\vspace{-4mm}
\end{center}
\end{table}

\begin{table}
\caption{ Ablation study on YouTube-VIS valid: individually integrating different components into the baseline.}
\vspace{-2mm}
\begin{center}
\setlength{\tabcolsep}{0.5mm}{
      \linespread{2}
      \begin{tabular}{p{0.095\textwidth}p{0.05\textwidth}<{\centering}
      p{0.045\textwidth}<{\centering}p{0.05\textwidth}<{\centering}p{0.05\textwidth}<{\centering}
      p{0.045\textwidth}<{\centering}p{0.05\textwidth}<{\centering}p{0.045\textwidth}<{\centering}}
        \Xhline{1pt}
         Methods         &     Impro.        &\ AP  &\ AP$_{50}$ &\ AP$_{75}$ &\ AP$_{S}$ &\ AP$_{M}$ &\ AP$_{L}$ \\
         \Xhline{0.5pt}
         Baseline        &      -           &28.9   & 45.8  & 30.1  & 9.8  & 26.9 & 40.3 \\
         + FCA           &    +1.9          &30.8   & 49.8  & 31.0  & 11.8 & 29.7 & 44.6 \\
         + FCB(ada)      &    +2.3          &31.2   & 52.1  & 31.2  & 11.1 & 32.8 & 43.6 \\
         + FCB(ali)      &    +2.4          &31.3   & 50.0  & 32.8  & 10.3 & 31.7 & 43.3 \\
         + TF            &    +5.2          &34.1   & 51.9  & 36.0  & 11.4 & 30.3 & 46.5 \\
         \Xhline{1pt}
      \end{tabular}\label{tab:ablation_VIS_single}
}
\vspace{-6mm}
\end{center}
\end{table}

\begin{table*}[!htb]
\caption{ Quantitative performance comparison on YouTube-VIS valid set, where a, b, c, and d respectively refer to semantic segmentation, high-solution mask refinement, temporal information, and multi-scale training.}
\vspace{-2mm}
\begin{center}
\setlength{\tabcolsep}{0.5mm}{
      \linespread{2}
      \begin{tabular}{p{0.09\textwidth}<{\centering}p{0.16\textwidth}p{0.11\textwidth}
      p{0.06\textwidth}<{\centering}p{0.07\textwidth}<{\centering}p{0.06\textwidth}<{\centering}
      p{0.05\textwidth}<{\centering}p{0.02\textwidth}
      p{0.055\textwidth}<{\centering}p{0.055\textwidth}<{\centering}p{0.055\textwidth}<{\centering}
      p{0.055\textwidth}<{\centering}p{0.055\textwidth}<{\centering}}
         \Xhline{1pt}
          Type&Methods &  Backbone & Frames & Anchors &Others & FPS  &  &\ AP  &\ AP$_{50}$ &\ AP$_{75}$ &\ AR$_{1}$ &\ AR$_{10}$\\
         \Xhline{0.5pt}
         \multirow{4}{*}{\shortstack{VOS \\ baseline}}
         &OSMN \cite{yang2018efficient}          &R50-FPN &3 &3 &-&-  & & 27.5 & 45.1 & 29.1  &28.6 &33.1\\
         &FEELVOS \cite{voigtlaender2019feelvos} &R50-FPN &3 &3 &-&-  & & 26.9 & 42.0 & 29.7  &29.9 &33.4\\
         &STEm-Seg \cite{athar2020stem}          &R50-FPN &- &- &-&-  & & 30.6 & 50.7 & 33.5  &31.6 &37.1\\
         &STEm-Seg \cite{athar2020stem}          &R101-FPN &- &- &-&2.1  & & 34.6 & 55.8 & 37.9  &34.4 &41.6\\
         \Xhline{0.3pt}
         \multirow{3}{*}{\shortstack{Two-stage \\ methods}}                 
         &MaskTrack \cite{yang2019video}\    &R50-FPN   & 2  &3 &- &18.4 & & 30.3 & 51.1 & 32.6 &31.0 &35.5 \\
         &MaskProp \cite{bertasius2020classifying}&R50-FPN   & 13 &3 &b,c,d &- & &  40.0 &  -   & 42.9 &- &-\\
         &MaskProp \cite{bertasius2020classifying}&R101-FPN   & 13 &3 &b,c,d &- & & 42.5 &  -   & 45.6 &- &-\\
         \Xhline{0.3pt}
         \multirow{5}{*}{\shortstack{One-stage \\ methods}} 
         &SipMask++ \cite{cao2020sipmask}        &R50-DCN & 2 &- & - & 30.0 & &  32.5 &  53.0 & 33.3  &33.5 &38.9 \\
         &SipMask++ \cite{cao2020sipmask}        &R101-DCN & 2 &- & - & 27.8 & &  35.0 &  56.1 & 35.2  &36.0 &41.2 \\
         &STMask (ada)                  &R50-DCN &2 &9 &c &28.6 & &33.5 &52.1 &36.9 &31.1 &39.2\\
         &STMask (ada)                  &R101-DCN &2 &3 &c&23.4 &&  36.8 &  56.8 & 38.0 & 34.8 &41.8 \\
         &STMask (ali)                  &R101-DCN &2 &3 &c&22.1 &&  36.3 &  55.2 &  39.9 &33.7 & 42.0 \\
         \Xhline{1pt}
      \end{tabular}\label{tab:VIS}
}
\vspace{-4mm}
\end{center}
\end{table*}

\begin{figure*}[!htb]
\begin{minipage}{0.49\linewidth}
  \centerline{\includegraphics[width=1.02\linewidth]{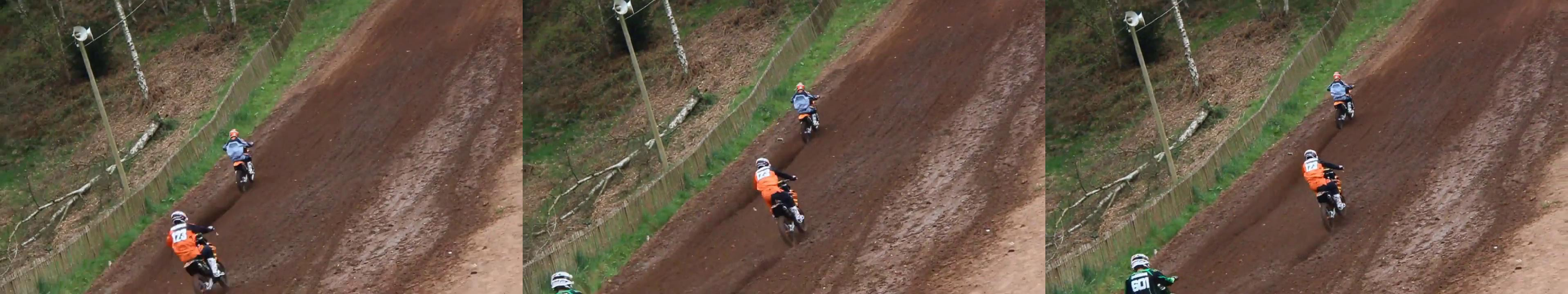}}
\end{minipage}
\hfill
\begin{minipage}{0.49\linewidth}
  \centerline{\includegraphics[width=1.02\linewidth]{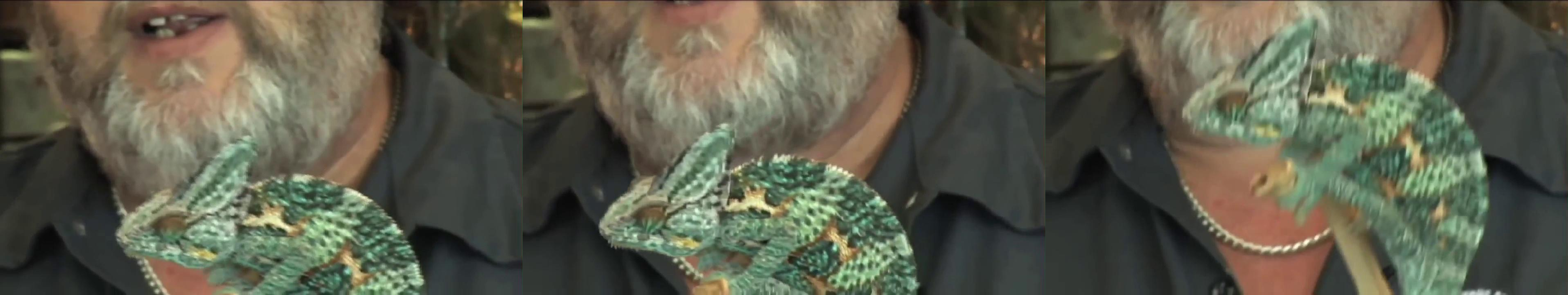}}
\end{minipage}
\vfill
\begin{minipage}{0.49\linewidth}
  \centerline{\includegraphics[width=1.02\linewidth]{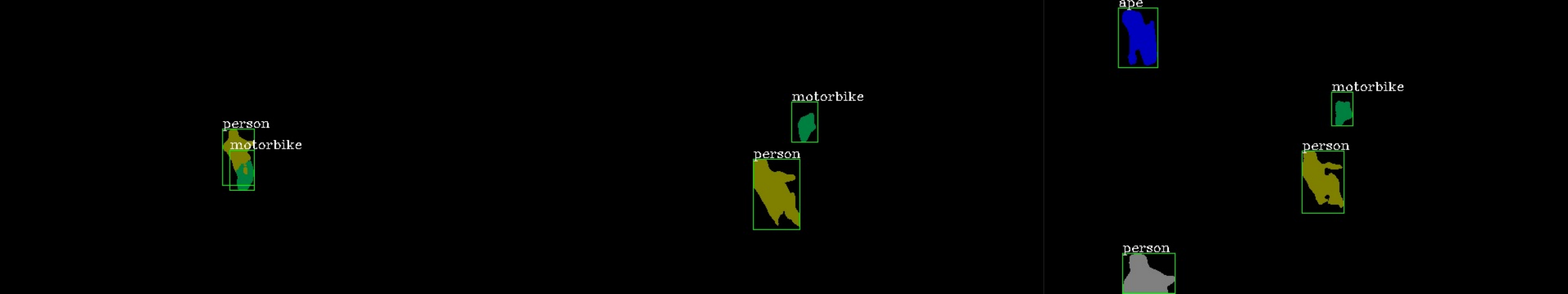}}
\end{minipage}
\hfill
\begin{minipage}{0.49\linewidth}
  \centerline{\includegraphics[width=1.02\linewidth]{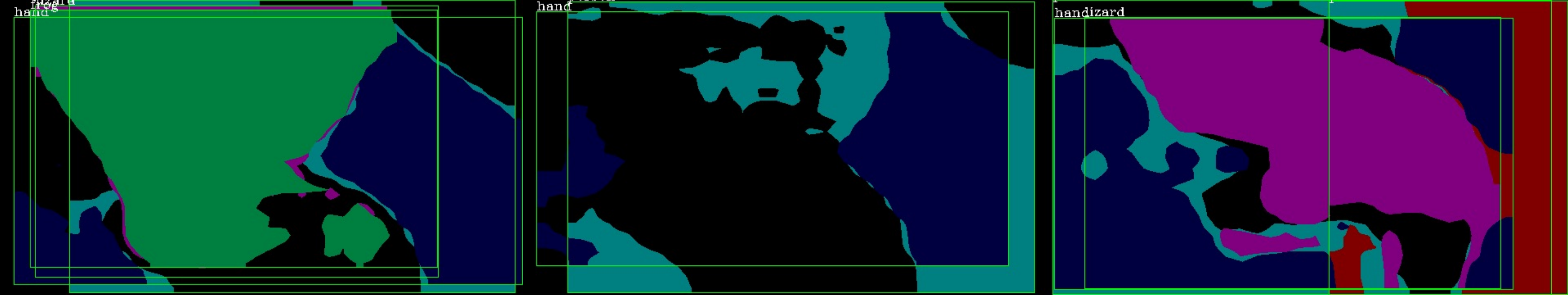}}
\end{minipage}
\vfill
\begin{minipage}{0.49\linewidth}
  \centerline{\includegraphics[width=1.02\linewidth]{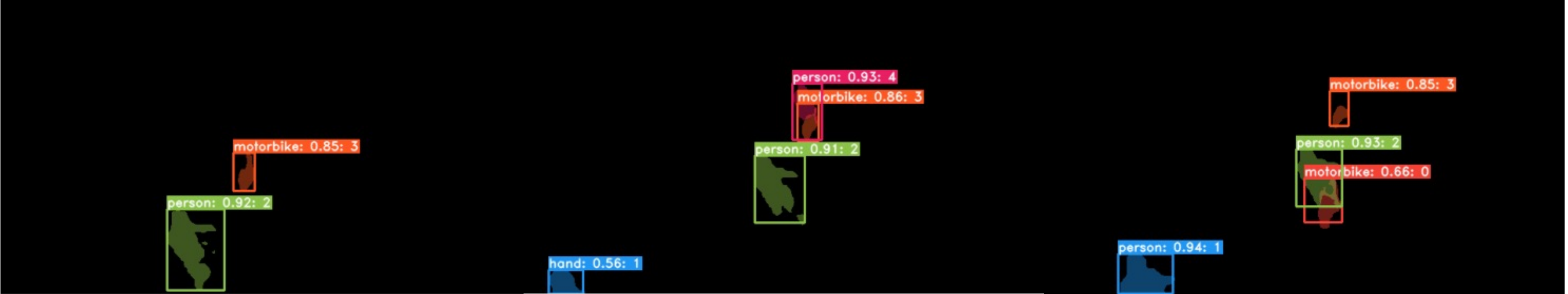}}
\end{minipage}
\hfill
\begin{minipage}{0.49\linewidth}
  \centerline{\includegraphics[width=1.02\linewidth]{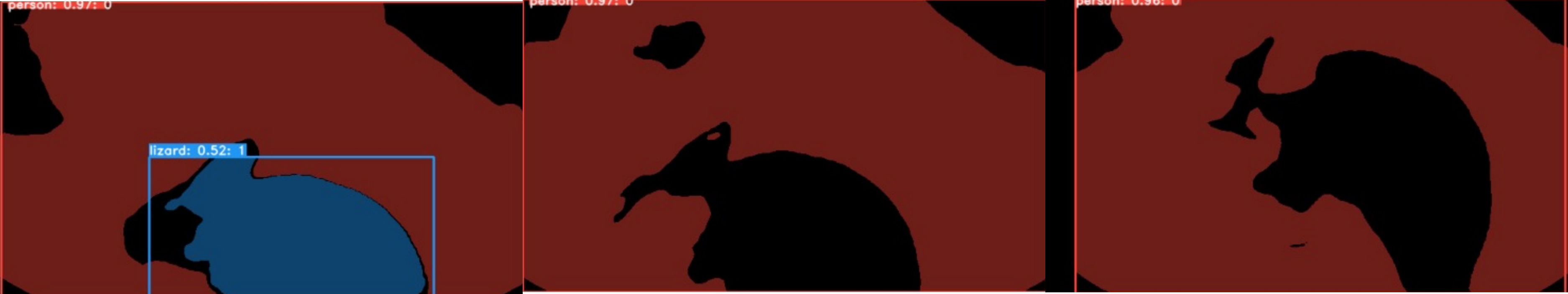}}
\end{minipage}
\vfill
\begin{minipage}{0.49\linewidth}
  \centerline{\includegraphics[width=1.02\linewidth]{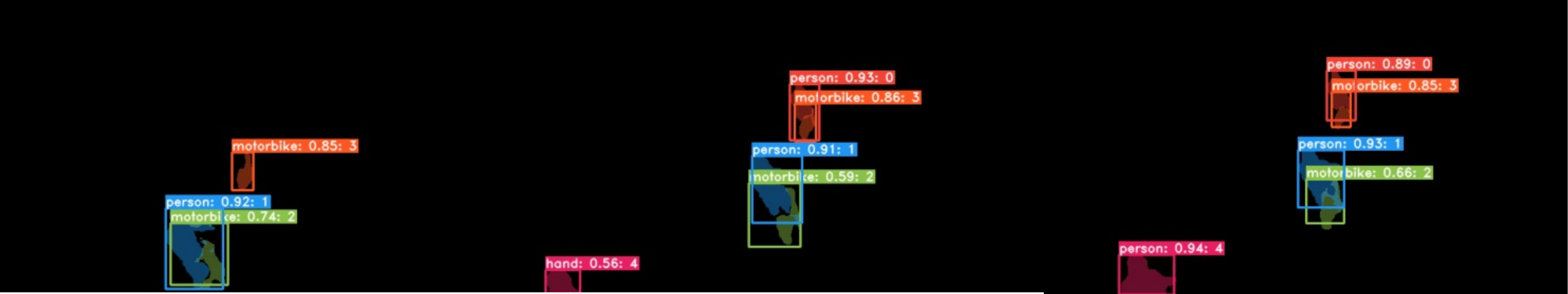}}
  \vspace{+1mm}
\end{minipage}
\hfill
\begin{minipage}{0.49\linewidth}
  \centerline{\includegraphics[width=1.02\linewidth]{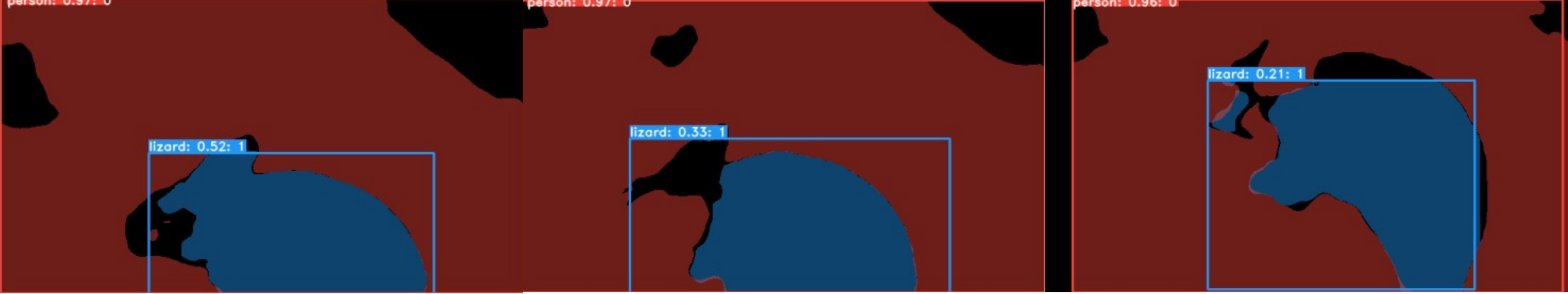}}
  \vspace{+1mm}
\end{minipage}
\caption{
Segmentation visual comparison on two videos with small objects and occlusion. The first two rows show the original frames and the instance masks produced by MaskTrack R-CNN \cite{yang2019video} respectively, while the last two rows are the results obtained by our STMask with only spatial calibration and with both spatial calibration and temporal fusion module, respectively.
}
\label{fig:visual1}
\vspace{-3mm}
\end{figure*}

\begin{figure}[!htb]
\begin{minipage}{0.99\linewidth}
  \centerline{\includegraphics[width=1.02\linewidth]{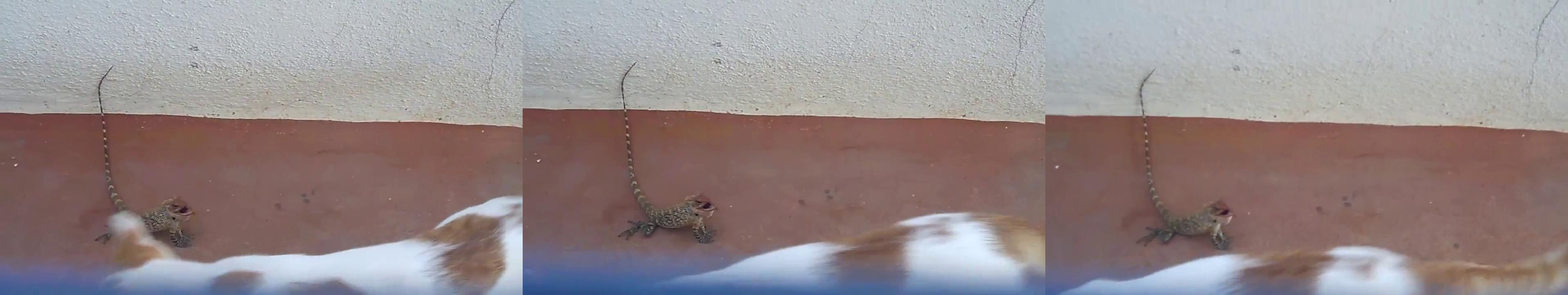}}
\end{minipage}
\vfill
\begin{minipage}{0.99\linewidth}
  \centerline{\includegraphics[width=1.02\linewidth]{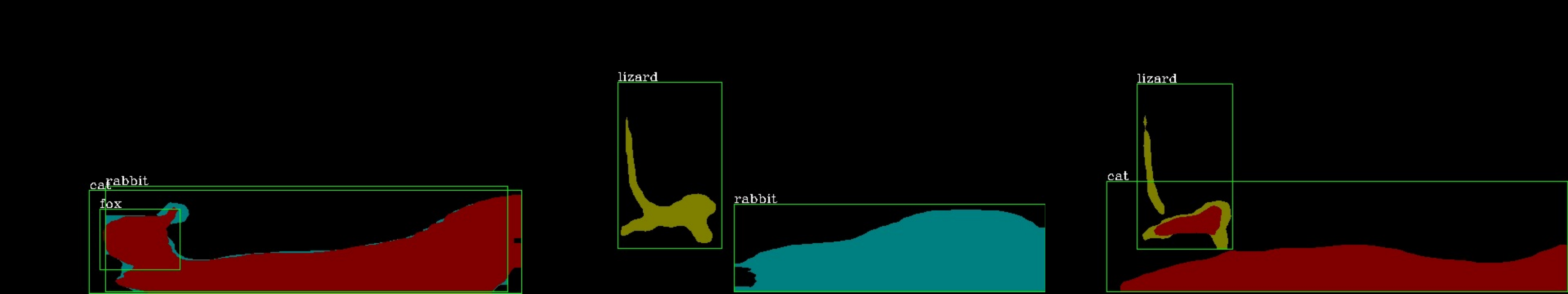}}
\end{minipage}
\vfill
\begin{minipage}{0.99\linewidth}
  \centerline{\includegraphics[width=1.022\linewidth]{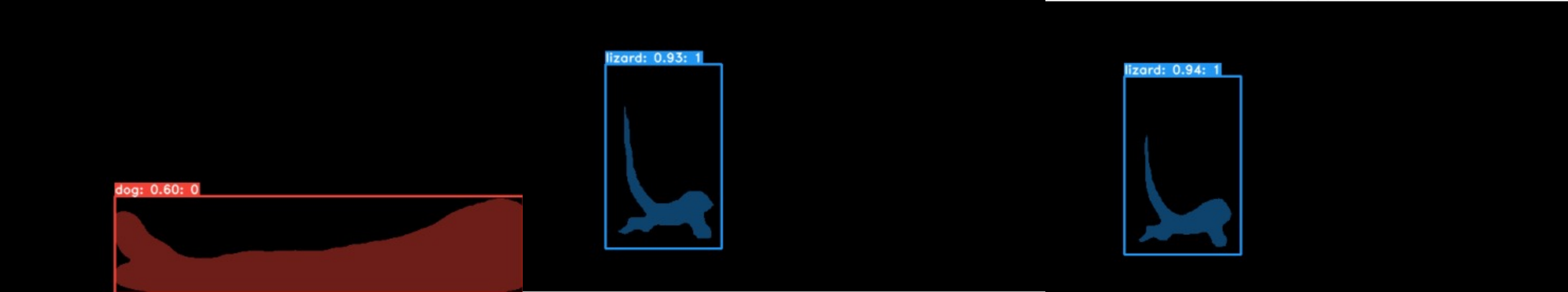}}
\end{minipage}
\vfill
\begin{minipage}{0.99\linewidth}
  \centerline{\includegraphics[width=1.02\linewidth]{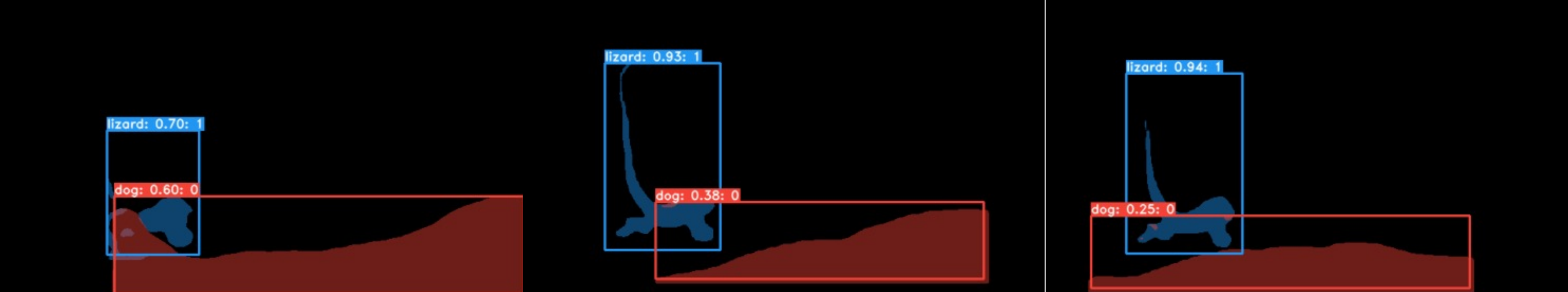}}
  \vspace{+1mm}
\end{minipage}
\caption{
Segmentation visual comparison on a video with uncommon camera-to-object view.
}
\label{fig:visual2}
\vspace{-6mm}
\end{figure}

\subsection{Mask Results}

We compare the proposed STMask with state-of-the-arts on YouTube-VIS valid set. The results are shown in Table \ref{tab:VIS}. The ResNet50 backbone is used by the competing methods. The speed is computed on a single 2080Ti GPU. 

From Table \ref{tab:VIS}, we can see that among existing fast video object segmentation (VOS) methods, 
OSMN \cite{yang2018efficient}  with `track-by-detect’ and FEELVOS \cite{voigtlaender2019feelvos} achieve relatively lower mask AP scores. 
The bottle-up method STEm-Seg with ResNet 50 and ResNet 101 backbone reaches 30.6 \% and 34.6 \% mask AP respectively.
The recently introduced two-stage methods MaskTrack R-CNN \cite{yang2019video} and MaskProp  \cite{bertasius2020classifying} obtain mask AP scores of 30.3 \% and 40.0 \%, respectively. 
However, MaskProp requires a video clip (more than 13 frames) to process an image, resulting in very slow speed. 
For the one-stage method using ResNet50 Backbone, SipMask \cite{cao2020sipmask} and our proposed STMask obtain 32.5 \% and 33.5 \% mask AP with a speed of 30.0 and 28.6 FPS respectively.
In addition, STMask with ResNet101-DCN backbone achieves 36.8 \%  and 36.3 \% mask AP on adaptive and aligned FCB settings, respectively, without any other extra tricks. Although STMask has relatively lower mask AP than MaskProp, it does achieve a better trade-off between accuracy and speed. 

Figs. \ref{fig:visual1} and \ref{fig:visual2} visualise the instance segmentation masks of our STMask on challenging videos with small objects, occlusion and uncommon camera-to-object view. We can see that the temporal fusion module can indeed reduce the missed detection across frames, compared with frame-level instance segmentation. 

\vspace{-1mm}
\section{Conclusion}
\vspace{-1mm}

We observe that one-stage instance segmentation approaches underestimate the importance of spatial feature calibration and temporal redundancy information between video frames for video instance segmentation.
To address the issue, we first propose a simple spatial feature calibration to detect and segment object masks frame-by-frame, and further introduce a temporal fusion module to track instance across video frames to effectively reduce missed instances on challenging videos like motion blur, partial occlusion and unusual object-to-camera poses.
Overall, on YouTube-VIS valid set, our proposed STMask with ResNet-50/-101 backbone obtains 33.5\% / 36.8 \% mask AP, while achieving 28.6 / 23.4 FPS on video instance segmentation, which does keep the trade-off between accuracy and speed.

\newpage
{\small
\bibliographystyle{ieee_fullname}
\bibliography{egbib}
}

\end{document}